\documentclass[letterpaper, 10 pt, conference]{ieeeconf}  
\IEEEoverridecommandlockouts                              

\usepackage{graphics} 
\usepackage{amsmath} 
\usepackage{amssymb}  
\usepackage{siunitx}
\usepackage{mathtools}
\usepackage{booktabs}
\usepackage[ruled]{algorithm2e}
\usepackage{subcaption} 
\usepackage[bottom=57pt,top=54pt, left=54pt, right=54pt]{geometry}   

\usepackage{adjustbox}
\usepackage{pbox}
\usepackage{dblfloatfix}
\usepackage{graphicx}
\usepackage[dvipsnames]{xcolor}
\usepackage{hyperref}
\usepackage{caption}
\usepackage{subcaption}
\usepackage[nospace]{cite}

\usepackage{array,multirow,graphicx}
\usepackage{makecell}

\DeclarePairedDelimiter\floor{\lfloor}{\rfloor}

\begin{document}

\title{
Panoptic Multi-TSDFs: a Flexible Representation for Online Multi-resolution Volumetric Mapping and Long-term Dynamic Scene Consistency
}


\author{Lukas Schmid$^{1}$, Jeffrey Delmerico$^2$, Johannes L. Schönberger$^2$, \\Juan Nieto$^2$, Marc Pollefeys$^{2,3}$, Roland Siegwart$^1$, and Cesar Cadena$^{1}$

\thanks{This work was supported by funding from the Microsoft Swiss Joint Research Center and the European Union’s Horizon 2020 research and innovation programme under grant agreement No 101017008.}
\thanks{$^1$ Autonomous Systems Lab, ETH Z\"urich, Z\"urich, Switzerland}%
\thanks{$^2$ Mixed Reality and AI Lab, Microsoft, Z\"urich, Switzerland}%
\thanks{$^3$ Computer Vision and Geometry Lab, ETH Zürich, Z\"urich, Switzerland}%
\thanks{
{\tt\small schmluk@ethz.ch}}%
}%

\maketitle

\begin{abstract}
For robotic interaction in environments shared with other agents, access to volumetric and semantic maps of the scene is crucial. However, such environments are inevitably subject to long-term changes, which the map needs to account for.
We thus propose \emph{panoptic multi-TSDFs} as a novel representation for multi-resolution volumetric mapping in changing environments. By leveraging high-level information for 3D reconstruction, our proposed system allocates high resolution only where needed. Through reasoning on the object level, semantic consistency over time is achieved. This enables our method to maintain up-to-date reconstructions with high accuracy while improving coverage by incorporating previous data. We show in thorough experimental evaluation that our map can be efficiently constructed, maintained, and queried during online operation, and that the presented approach can operate robustly on real depth sensors using non-optimized panoptic segmentation as input. 
\end{abstract}

\section{Introduction}
\label{sec:introduction}

Having a geometric and semantic understanding of the world is a crucial capability for autonomous systems to interact with their environment in tasks ranging from collision avoidance and path planning to mobile manipulation or object search. In many applications, these tasks are desirable in environments that are shared with other agents. However, these inevitably induce long-term dynamic changes in the environment that the robot map needs to account for.

In particular, volumetric representations such as occupancy~\cite{hornung2013octomap} or Truncated Signed Distance Fields (TSDF)~\cite{oleynikova2017voxblox} have found a lot of success. By dividing the map into a dense grid of voxels, they are able to explicitly represent free space and differentiate between known and unknown regions in the map, which is crucial for online planning. However, this fixed grid structure makes these methods very memory intensive and renders them inflexible when trying to account temporal changes.

\begin{figure}
    \centering
    \begin{subfigure}{.493\linewidth}
      \centering
      \includegraphics[width=\linewidth]{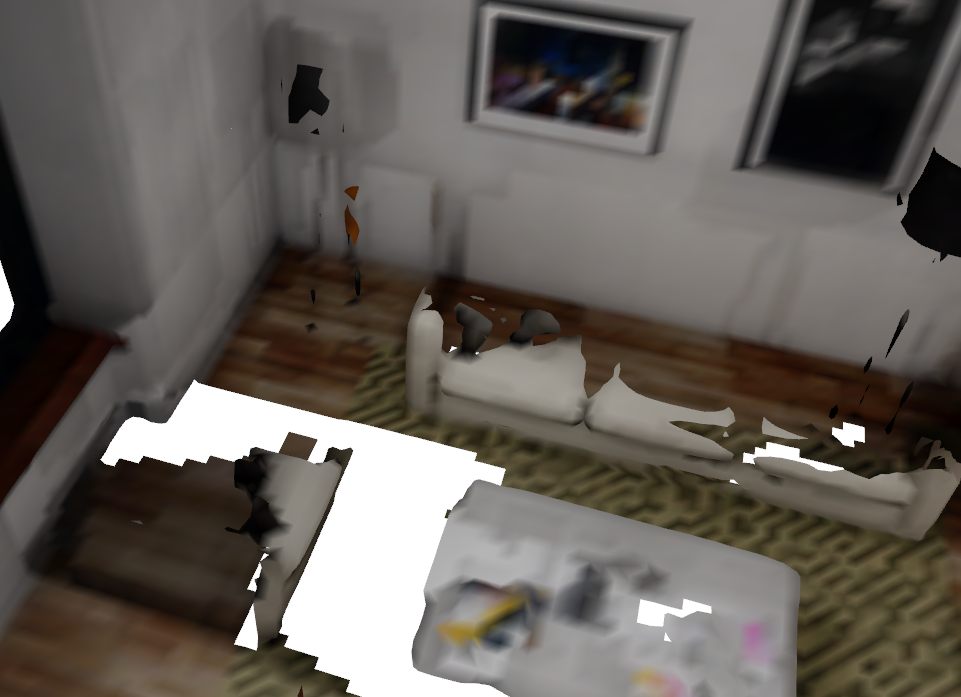}
      \caption{Monolithic map.}
    \end{subfigure}%
    \hspace{0.1mm}
    \begin{subfigure}{.493\linewidth}
      \centering
      \includegraphics[width=\linewidth]{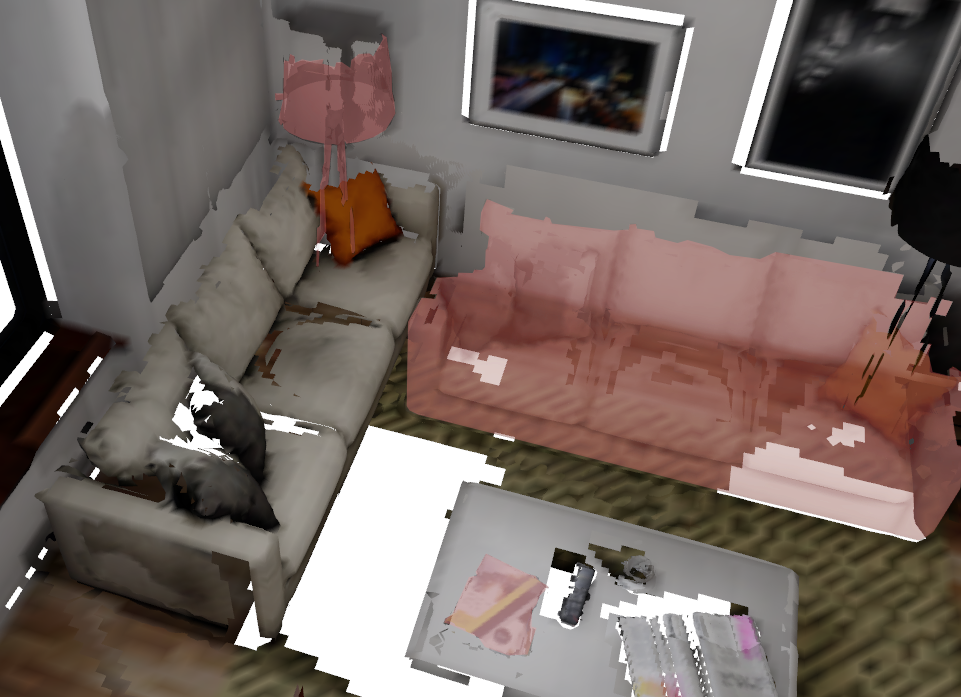}
      \caption{Panoptic multi-TSDFs (Ours).}
    \end{subfigure}%
    \vspace{-4pt}
    \caption{Qualitative comparison. Our method shows persistent and absent meshes in solid and shaded red, respectively. Our object-oriented approach preserves semantic consistency over time, accurately capturing new objects (left sofa) and removing absent objects (right sofa) as a whole. In contrast, the voxel-based approach keeps artifacts in the map and fails to capture up-to-date geometry. In addition, individual objects (on the table) are not merged together and can be reconstructed at higher resolutions.}
    \label{fig:sem_consistency}
    \vspace{-21pt}
\end{figure}

Recently, a number of works extended dense maps to also provide semantic information~\cite{panoptic_fusion, semantic_fusion, strecke_em_fusion_2019, co_fusion,runz_maskfusion_2018, grinvald_volumetric_2019, rosinol_kimera_2020}. Typically, semantic image predictions obtained by Convolutional Neural Networks (CNN) are fused into a global map to estimate the maximum a posteriori labels for each voxel. However, these methods still assume that the environment is static in order to integrate semantics into the fixed geometry.

In this work, we aim to invert this paradigm and explore how semantic information can be leveraged to improve the modeling of geometry and achieve temporal consistency. The central idea of our approach is that the world typically does not change at random but in a semantically consistent way. We thus propose a novel semantic volumetric map representation that uses the object as the minimal unit of change, rather than the voxel. Based on the panoptic segmentation paradigm~\cite{kirillov2019panoptic}, we differentiate between object instances, background classes, and free space. In light of recent success of submapping approaches for spatially consistent volumetric mapping~\cite{reijgwart2019voxgraph, schmid2021unified} and moving object reconstruction~\cite{strecke_em_fusion_2019, co_fusion, xu_mid_fusion}, we represent the world as a collection of submaps. Each submap contains a locally consistent panoptic entity, i.e. each object, piece of background, or free space is reconstructed individually, such that the collection of submaps together recovers the full volumetric map. 
We show that this panoptic map representation enables memory efficient multi-resolution volumetric mapping and is able to capture long-term dynamic scene changes during online mapping.
We make the following contributions:
\begin{itemize}
    \item We propose panoptic multi-TSDFs as a novel, flexible multi-resolution volumetric map representation to capture long-term object-level scene changes.
    \item We present a method for panoptic multi-TSDF integration and map management for temporally consistent mapping during online operation.
    \item We thoroughly evaluate our approach in simulation and on real world datasets. The code and data is available as open-source\footnote{\url{https://github.com/ethz-asl/panoptic_mapping}}.
\end{itemize}

\section{Related Work} 
\label{sec:rel_work}

\subsection{Dense Semantic Mapping}
\label{sec:rel_dense}

Dense semantic mapping aims at estimating the semantic label of each surface element. 
Early works~\cite{tateno_real-time_2015} fuse frame-wise geometric segmentations into a global surfel-map.
McCormac~\textit{et~al.}~\cite{semantic_fusion} extend surfel-based mapping~\cite{whelan2015elasticfusion} by fusing 2D semantic predictions and refining using a global Conditional Random Field (CRF).

Similarly,~\cite{rosinol_kimera_2020} fuse CNN predictions in a Bayesian way into a volumetric map based on voxblox~\cite{oleynikova2017voxblox}. Grinvald~\textit{et~al.}~\cite{grinvald_volumetric_2019} extend \cite{oleynikova2017voxblox} by combining geometric segmentation with instance predictions from MaskRCNN~\cite{he2017maskrcnn} to refine label boundaries. A panoptic approach also based on \cite{oleynikova2017voxblox} is presented in~\cite{panoptic_fusion}, where CNN background labels are combined with instance predictions~\cite{he2017maskrcnn} to achieve the panoptic labeling.
While the TSDF-based methods can supply the volumetric information needed for planning, all of the above approaches make the limiting assumption that the environment is static.

\subsection{Object-centric Mapping}
\label{sec:rel_objects}

A different family of approaches focuses on reconstructing selected individual objects.
This was pioneered in SLAM++~\cite{salas2013slampp}, where models of known objects are fitted to sensor data and act as nodes in graph-based sparse SLAM. This constraint is relaxed in Fusion++~\cite{mccormac_fusion_2018}. Similar to us, each object is reconstructed in its own TSDF volume and segmented by estimating a foreground probability, giving flexibility to the system to account for pose estimation errors. However, only selected objects are reconstructed, thus not providing the volumetric information required for planning. Furthermore, the environment is considered static.

A number of works leverage this approach to capture short-term dynamics, i.e. objects moving in front of the camera. Rünz~\textit{et~al.}~\cite{co_fusion} track objects using geometric and photometric alginment. 
They are segmented based on motion cues or semantic segmentation and reconstructed using surfel fusion. In a similar approach, MaskFusion~\cite{runz_maskfusion_2018} combines geometric and instance segmentation~\cite{he2017maskrcnn} for improved object recognition.
Strecke~\textit{et~al.}~\cite{strecke_em_fusion_2019} extend \cite{mccormac_fusion_2018} to moving objects, estimating camera and object poses in an expectation-maximization scheme. In a TSDF approach based on \cite{vespa_supereight}, MID-Fusion~\cite{xu_mid_fusion} combines the segmentation of \cite{runz_maskfusion_2018} with motion cues to reconstruct multiple moving objects. Long~\textit{et~al.}~\cite{long_rigidfusion_2021} additionally include motion tracking to reconstruct a single large moving object.

While significant progress in reconstructing selected individual objects was made, these approaches are usually confined to small environments with few tracked objects. Non-moving objects and background are not considered and assumed static, thus making these approaches not well suited to capture long-term changes. 

\subsection{Change Detection}
\label{sec:rel_change}

TSDFs have also found success in capturing long-term changes. Finman~\textit{et~al.}~\cite{finman_2013_lifelong} generate multiple reconstructions \cite{whelan2012kintinuous} and identify changes via surface point cloud differencing. Fehr~\textit{et~al.}~\cite{tsdf_change} directly combine different observations into a multi-layer TSDF and use volumetric differencing to extract the static background and movable objects. A recent approach~\cite{langer2020robust} additionally leverages semantic information to identify support surfaces and movable objects to improve change detection. However, all of these methods can only operate post-hoc and are computationally demanding, making them unsuitable for online operation of robots in shared environments.

\subsection{Online Long-term Mapping}
\label{sec:rel_longterm}

Another line of works tackles the problem of incremental long-term mapping. 
Krajník~\textit{et~al.}~\cite{frequency_map} augment 2D occupancy maps to estimate the temporal presence of each voxel as a frequency. Lázaro~\textit{et~al.}~\cite{longterm_pointclouds} represent the world as 2D point cloud submaps and apply a map management strategy, similar measurements are fused and differing data is overwritten by the most recent estimate.
Tang~\textit{et~al.}~\cite{tang2019topological} build a graph of submaps connected by their poses. When new submaps can not be re-localized, they are also added as temporal information to the spatial graph. 
Alternatively, Macenski~\textit{et~al.}~\cite{macenski2020spatio} add a temporal decay to voxels. As old voxels are removed the map is kept up to date but also loses previous information.
Mason and Marthi~\cite{mason2012object} propose an object-based approach, where an object is any point cloud supported by a plane. Persistence of these sparse objects is then tracked by comparing their convex hulls projected onto the support plane.

A limitation of these approaches is that they lack the expressiveness, i.e. volumetric and semantic information, needed for robot interaction and do not provide semantic consistency when accounting for changes, thus leaving undesirable artifacts in the map. 

\section{Approach} 
\label{sec:approach}

\begin{figure}
    \centering
    \includegraphics[width=\columnwidth]{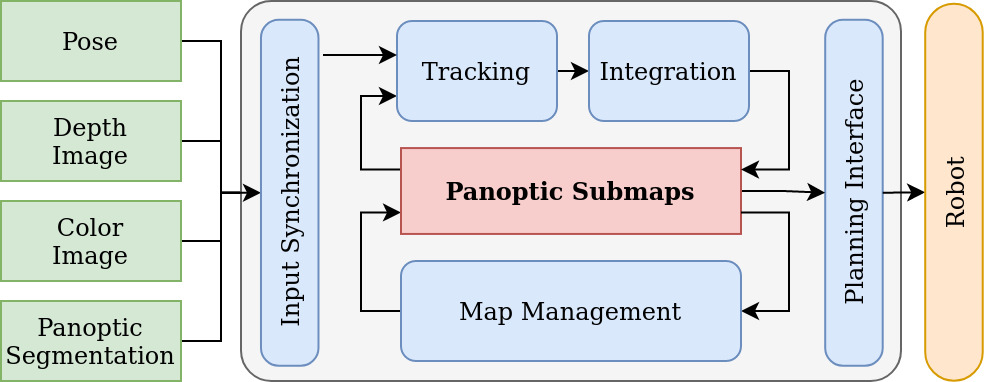}
    \caption{System overview. 
    }
    \label{fig:system}
    \vspace{-24pt}
\end{figure}

The central idea of our approach is to leverage panoptic segmentation information as the governing factor in representing, building, and maintaining volumetric maps during online operation. 
The goal of our work is not to optimize the semantic labeling, but rather to explore how high-level information can be leveraged for multi-resolution 3D reconstruction and temporal consistency.
An overview of our system is given in Fig.~\ref{fig:system}.
The inputs to our pipeline are depth and color images, e.g. from a
RGBD sensor. We take robot poses from an external estimator, allowing for a broad range of sensors and systems, e.g.  \cite{izadi2011kinectfusion, bloesch2017iterated, zuo2019lic}, to be employed. 
Lastly, panoptic segmentation can be predicted from the color and depth information. To demonstrate the robustness of our method with respect to imperfect segmentation, we directly use the output of \cite{wu2019detectron2} as input to our system. Nonetheless, our method can readily integrate other segmentation improvements such as \cite{semantic_fusion, panoptic_fusion, tateno_real-time_2015,grinvald_volumetric_2019, runz_maskfusion_2018, xu_mid_fusion}.

\subsection{Map Representation}
\label{sec:app_representation}

Our map representation is based on the observation that the world typically does not change at random but in a semantically consistent way. To capture this feature, we use the object as the minimal unit of change and propose to represent the world as a collection of panoptic entities, structured as submaps. In this formulation, we differentiate between three panoptic labels, being \emph{objects}, \emph{background}, and \emph{free space}. Each submap contains the geometry of one entity, i.e. of an object instance, a background class, or free space, such that all submaps together constitute the full volumetric map. To guarantee temporal consistency of each submap, we further differentiate between \emph{active} and \emph{inactive} submaps, denoting active those currently being tracked and built, and inactive submaps from past observations.

\begin{figure}
    \centering
    \includegraphics[width=1.0\columnwidth]{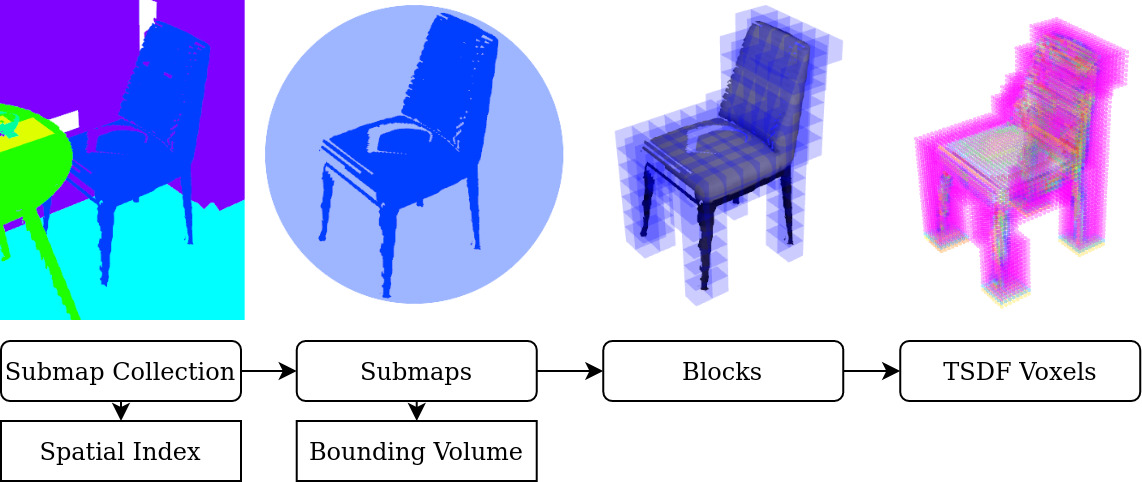}
    \caption{Hierarchical map representation. Each submap (color) contains locks of TSDF voxel grids (colored pink to green based on the TSDF values). Submap bounding volumes and a global spatial index are maintained for efficient map traversal.}
    \label{fig:map}
    \vspace{-18pt}
\end{figure}

For efficient processing, a hierarchical structure illustrated in Fig.~\ref{fig:map} is employed. On the highest level lies the submap collection, where a spatial index 
is maintained for constant time scaling in large-scale environments. 
Each submap contains all related data such as panoptic, instance, and class labels, as well as transformation and tracking data. To represent geometry, we choose to use TSDF grids \cite{oleynikova2017voxblox} for their ability to fuse multiple observations. The space containing an object is partitioned into blocks, where only blocks containing surface information are allocated, except for free space submaps. For efficient traversal of the submap collection, each object has a sphere spanned by the blocks as bounding volume. Each block contains a dense grid of voxels that store the TSDF values representing the surface.

This hierarchical structure allows for efficient queries of the submap collection at all stages of the pipeline. In addition, each object can be reconstructed at a different resolution and only takes up the memory required to represent its surface, while the full volumetric information can be recovered from the collection. Most importantly, semantic consistency is maintained by performing reasoning, e.g. about persistence over time, on the object level. This further makes our approach very flexible to also account e.g. for short-term dynamics via object tracking \cite{co_fusion, runz_maskfusion_2018, strecke_em_fusion_2019, xu_mid_fusion} or global consistency \cite{reijgwart2019voxgraph, schmid2021unified}. However, this is left for future work. Finally, because all submaps are fully data-parallel, the following operations can be distributed over multiple cores.

\subsection{Label Tracking}
\label{sec:app_tracking}

To ensure the consistency of instance labels over multiple frames and temporal consistency of the map, incoming frames are tracked against the current map. Since ray-casting into many submaps quickly becomes intractable, we incrementally compute the iso-surface of each submap using Marching Cubes \cite{lorensen1987marching}. All active submaps whose bounding volume intersects with the view frustum are gathered and iso-surface points of blocks in the view frustum are projected into the image plane. Points whose rendered depth lies within a tolerance of $\xi_{d} = \nu$, where $\nu$ denotes the voxel size of the TSDF, of the measured depth are considered valid and fill in a patch of size $\nu$.
Each input segment is associated to the submap that has the highest Intersection over Union (IoU) between predicted and rendered mask and the same class label. To avoid spurious associations, a minimum IoU of $\xi_{IoU} = 0.1$ is required.

For masks that were not associated, a new submap is allocated. Compared to other approaches \cite{mccormac_fusion_2018, strecke_em_fusion_2019, xu_mid_fusion} that allocate a fixed grid size such that the object is contained in it, we choose $\nu$ as a function of the semantic label of each detection. This allows to select $\nu$ e.g. between $[\nu_{small}\in\{2\dots4\}, \nu_{large} \in\{ 5\dots 10\}]$ cm based on how complex the expected geometry of the object class is and leave the efficiency optimization to our hierarchical map structure. We keep $\nu_{freespace}=30$ cm for all settings. To guarantee local consistency, submaps are only active as long as they are successfully tracked, leading to multiple submaps with potentially varying resolutions describing the same object in case of detection or tracking failure. These submaps are later filtered during map management. To avoid instantiating too many false positives, submaps need to be tracked for $\tau_{new}=3$ frames to be kept. Similarly, submaps that have not been detected for $\tau_{active}=5$ frames are deactivated. This ensures that data is only integrated into currently tracked submaps and previous data can not be corrupted when e.g. changes in the environment have occurred.

\subsection{Integration}
\label{sec:app_integration}

To update the volumetric map, each measurement is fully integrated into all active submaps. Since ray-casting as in \cite{oleynikova2017voxblox} quickly becomes intractable for many submaps, we use our hierarchical map representation for fast projective updates. For each submap, all blocks within the truncation distance $\delta=2\nu$ of points belonging to their masks are allocated. Similar to \cite{mccormac_fusion_2018}, we separately reconstruct geometry and semantics. To best estimate the surfaces, we perform TSDF updates to \emph{all} allocated voxels, adapting the weighting function of \cite{oleynikova2017voxblox}:

\vspace{-15pt}
\begin{equation}
    w_{in}(v) = \frac{f_x * f_y * \nu^2}{z(v)^4} 
\end{equation}

Where $f_x$ and $f_y$ are the focal lengths of the camera and $z(v)$ is the depth of voxel $v$ in the image. 

To refine which surfaces are part of the submap, each voxel $v$ has a belonging probability $P_b(v)$. Since network probabilities can be overconfident \cite{mccormac_fusion_2018, strecke_em_fusion_2019}, 
we employ a memory efficient binary estimate $P_b(v)$ of the count probability $P_b^\star(v)$ using weights $p$:

\vspace{-10pt}
\begin{gather}
    P_b^\star(v) = 
    \frac{1}{|\mathbb{F}_t|}
    \sum_{f = 1}^{|\mathbb{F}_t|} \mathbb{I}_{\{label(u_f(v)) = label(S(v)) \}}
    \label{eq:p_belongs_true}
\\
    P_b(v) = 
    \frac{
    \sum_{f = 1}^{|\mathbb{F}_t|} p(|\mathbb{F}_t|-f) \mathbb{I}_{\{label(u_f(v)) = label(S(v)) \}}
    }{
    \sum_{f = 1}^{|\mathbb{F}_t|} p(|\mathbb{F}_t|-f)
    }
    \label{eq:p_belongs}
\\
    p(f) = 1/2^{\floor*{f/128}}
    \label{eq:p_weight}
\end{gather}

where $\mathbb{F}_t$ is the set of frames where submap $S(v)$ was tracked, $u_f(v)$ is $v$ projected into image $f$, and $\mathbb{I}$ is the indicator function. This way, $P_b(v)$ can always be efficiently stored in only 16 bits.

Blocks that do not contain relevant information, i.e. $\nexists$ voxel $v$ s.t. $|sdf(v)| < \delta \wedge P_b(v) > 0.5$, are pruned.

\subsection{Map Management}
\label{sec:app_management}
Inactive submaps are frozen except for their change state $C(S) \in \{\text{persistent, unobserved, absent}\}$.
To compare two submaps, we interpolate the SDF distance $sdf(p)$ and weight $w(p)$ of each iso-surface point $p\in \mathbb{P}$ of the reference map in the other map. For each observed point, the distance should be close to 0 if the point is a surface. If $|sdf(p)| < \xi_{sdf}$, where $\xi_{sdf} = \nu$ is the error tolerance, the point counts as agreeing with the surface. Otherwise, $sdf(p) < -\xi_{sdf}$ indicates intersections with object maps and $sdf(p) > \xi_{sdf}$ indicates conflicts with free space maps. Each point is weighted with the combined TSDF weight:

\vspace{-10pt}
\begin{equation}
    \hat{w}(p) = \sqrt{\min\left(\frac{w(p)}{\xi_w}, 1\right) * \min\left(\frac{w_{ref}(p)}{\xi_w}, 1\right)}
    \label{eq:combined_weight}
\end{equation}

We empirically set the max weight $\xi_w=100$. Submaps count as conflicting or matching if the weight-adjusted number of points exceeds a threshold $\tau_{abs}=20$ or $\tau_{rel}=2\%$ of $|\mathbb{P}|$.

When performing change detection, all inactive submaps that overlap with active submaps are compared against the latter. If they conflict with any of them, their state is set to \emph{absent}. Otherwise, if they match with any of them, their state is set to \emph{persistent}. This way, erroneous matches or rejections, e.g. through sensing noise, are corrected for later on. Far back in time submaps are \emph{unknown}, and can become absent or persistent again when observed.

When submaps are deactivated and match with inactive submaps of the same class, they are fused together, allowing re-use of prior measurements and connections of components separated by e.g. occlusions or missed detections.

\subsection{Map Queries}
\label{sec:app_queries}
To utilize the map for robotic interaction, efficient queries
are important. To achieve this, we make use of our hierarchical map representation to only consider submaps and blocks that intersect with a query point $p$. Similarly, we use a temporal hierarchy to query spatio-temporal information. If the point is observed in an active submap, we directly use the highest resolution submap. Otherwise, $sdf(p)$ is the minimum of the distances to the surface of any persistent submap. Lastly, free space submaps are queried before resorting to yet unobserved submaps to predict expectations. Only present, i.e. \emph{active} or \emph{persistent} submaps, are counted for evaluation.


\section{Evaluations}
\label{sec:results}

\begin{figure}
    \centering
    \includegraphics[width=0.32\columnwidth]{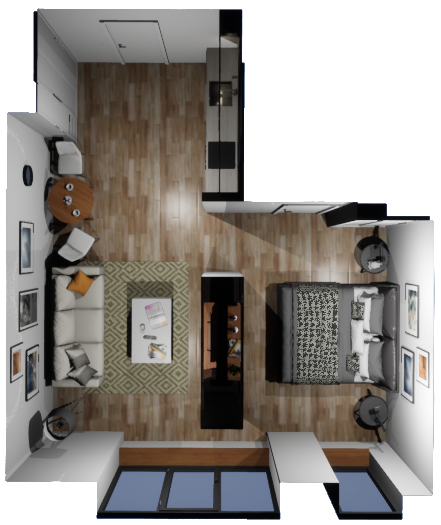}
    \includegraphics[width=0.33\columnwidth]{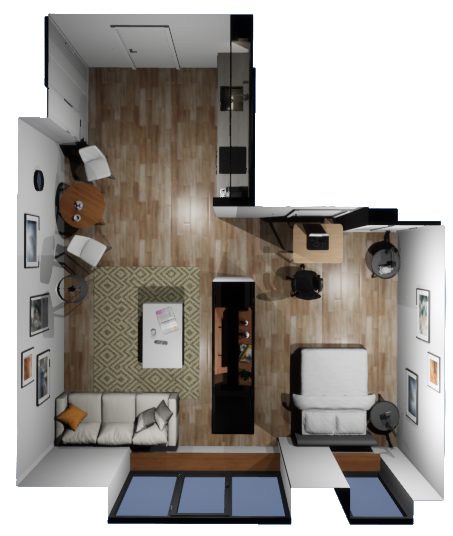}
    \includegraphics[width=0.32\columnwidth]{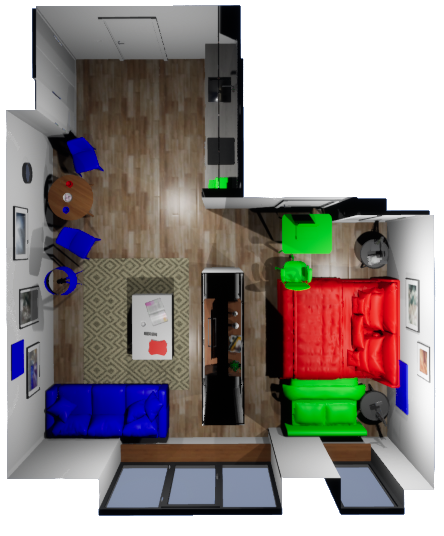}
    \caption{Flat dataset. Run 1 (left), run 2 (center), and changes (right), showing new (green), removed (red), and modified (blue) objects.}
    \label{fig:flat_data}
    \vspace{-18pt}
\end{figure}

To properly evaluate temporal consistency, the true geometry of the whole scene needs to be known at every time step. Since this is hard to obtain in the real world, we employ a simulated environment, where complete ground truth is available. This data, termed the \emph{flat dataset}, consists of two trajectories in a flat, where 8 objects are moved, 5 are added and 4 are removed between the runs, highlighted in in Fig.~\ref{fig:flat_data}. The data was generated using the high fidelity simulation of \cite{schmid2021unified}.
We make this dataset available for future comparisons. To verify our method using real sensors and scenes, we perform experiments on the RIO dataset \cite{Wald2019RIO} with the limited ground truth available. We use the provided ground truth or optimized poses for state estimation.

\subsection{Multi-resolution}
\label{sec:res_multires}

\begin{figure}
    \centering
    \includegraphics[width=1.0\columnwidth]{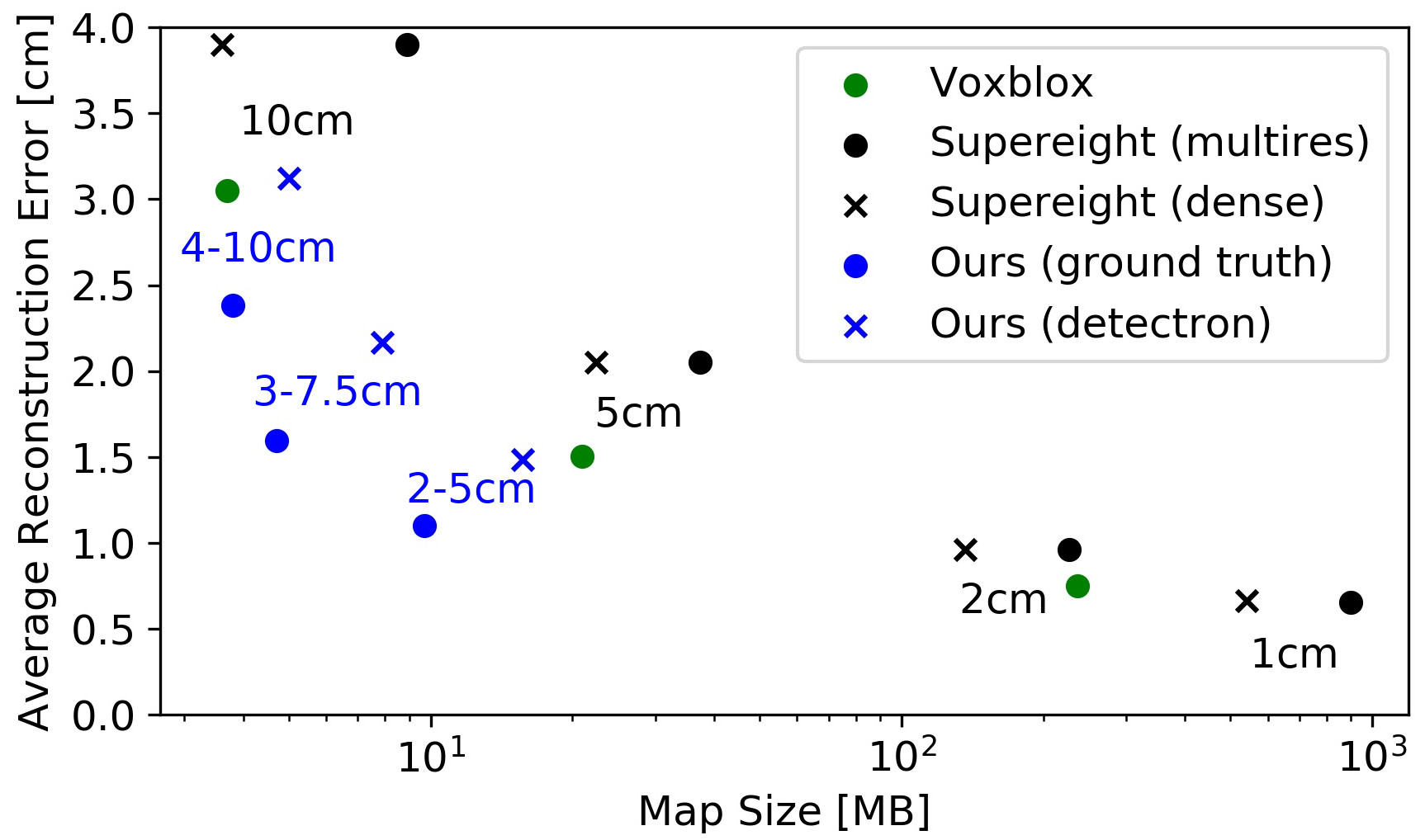}
    \vspace{-15pt}
    \caption{Error vs map size for different voxel sizes indicated as text. For Supereight (multires) the minimum voxel size is given. 
    }
    \label{fig:multires}
    \vspace{-14pt}
\end{figure}

Fig.~\ref{fig:multires} shows the reconstruction error as Mean Absolute Distance (MAD) versus the map size for varying voxel sizes, evaluated in the flat dataset. We compare against Voxblox~\cite{oleynikova2017voxblox} and Supereight\footnote{We thank Emanuele Vespa for support and discussion while setting up Supereight.}~\cite{vespa_supereight}, which are the geometry representations of many semantic mapping frameworks \cite{grinvald_volumetric_2019, panoptic_fusion, rosinol_kimera_2020, xu_mid_fusion}. We further compare against the sensor-based multi-resolution approach of \cite{vespa2019adaptive}.

The use of Ground Truth (GT) segmentation highlights the potential capabilities of our method, almost cutting the reconstruction error in half while consuming similar memory to Supereight for low resolutions, and achieving similar quality to \cite{vespa2019adaptive} while reducing the memory $\times$23 for high resolutions. 
This suggests significant benefits of semantically informed multi-resolution over the purely geometry-based approach.
Even with naive segmentation inputs, where only few small objects are detected and reconstructed precisely and false detections increase the memory consumption, our hierarchical map still saves memory compared to the baselines.

\subsection{Long-term Temporal Consistency}
\label{sec:res_longterm}

\begin{figure}
    \centering
    \includegraphics[width=1.0\columnwidth]{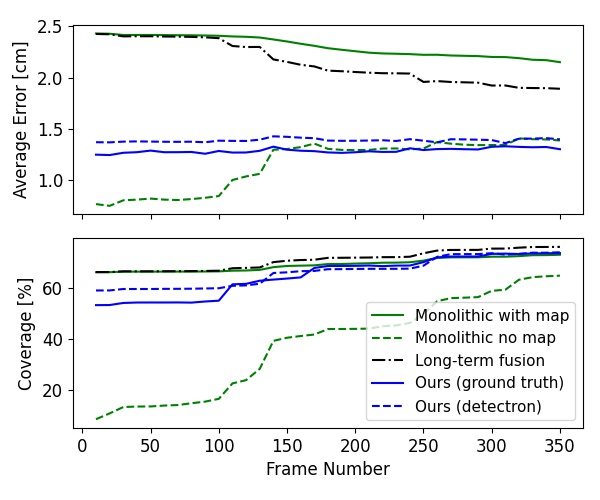}
    \vspace{-25pt}
    \caption{Long-term mapping performance shown for the second run in the flat dataset, starting from a map of the first. 
    }
    \label{fig:longterm}
    \vspace{-20pt}
\end{figure}

Fig.~\ref{fig:longterm} shows the MAD error and coverage as percentage of ground truth points with an error  $<$5 cm during the second trajectory of the flat dataset, given a prior map from the first run. We compare against a monolithic map as in \cite{grinvald_volumetric_2019, rosinol_kimera_2020, panoptic_fusion} and consider both continueing mapping based on the previous state, or starting from scratch. This separation is only done for comparison, as our system accounts for changes continuously. We further compare against our implementation of \cite{longterm_pointclouds} for volumetric maps, labeled Long-term fusion, where voxels that have a distance update $>$5 cm are overwritten. For fair comparisons that only focus on the temporal component, we use identical TSDF integration and constant $\nu=5$ cm for all approaches.

By including only active or persistent submaps, the reconstruction accuracy of our approach remains consistently at the discretization accuracy of $\sim1.4$ cm. The map built from scratch initially shows a lower error, since the second run starts in a kitchen with few complex surfaces.
When starting from the previous map, none of the changes are captured, resulting in high errors. As more observations are made, the error is reduced. This happens faster when for long-term fusion strategy. However, due to the lack of semantic consistency, artifacts in the map keep the error well above the discretization level. Through inclusion of previous data that match current observations, our method explores  significantly faster, converging to full coverage.
The small gap between GT and Detectron highlights the robustness of our method to imperfect segmentation, since it only assumes semantic consistency of each submap and does not require completeness or accuracy.

\subsection{Semantic Consistency}
Qualitative comparisons are shown in Fig.~\ref{fig:sem_consistency}. Since in our formulation semantically consistent submaps are the minimal unit of change, object consistency is preserved over time. In comparison, the voxel-based approach results in artifacts in the map and objects being merged together.

\subsection{Spatio-temporal Look-ups for Online Planning}

\begin{figure}
    \centering
    \includegraphics[width=1.0\columnwidth]{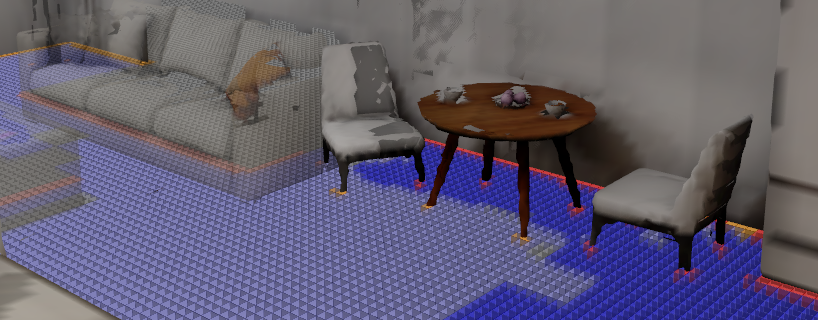}
    \caption{Spatio-temporal map queries. Meshes of present submaps are drawn in solid and points observed as occupied or free are shown in solid red and blue, respectively. Yet unobserved points are shown in orange if they are occupied by persistent objects, or in shaded red and blue if they are expected to be occupied or free based on previous data (shaded meshes). Unknown points are gray.}
    \label{fig:map_lookups}
    \vspace{-20pt}
\end{figure}

Fig.~\ref{fig:map_lookups} shows spatio-temporal occupancy look-ups on our proposed map representation. Although our map can be queried at any point in space, a 2D slice is visualized. Fig.~\ref{fig:map_lookups} highlights both the spatio-temporal information retrieved from single map queries, as well as multi-resolution preserving thin geometry. Map look-ups on this collection consisting of 130 submaps took an average of $3\mu s$.

\subsection{Real World Experiments}
\label{sec:res_rio}
\begin{figure*}[b!]
\centering
\begin{subfigure}{.25\textwidth}
  \centering
  \includegraphics[width=\linewidth]{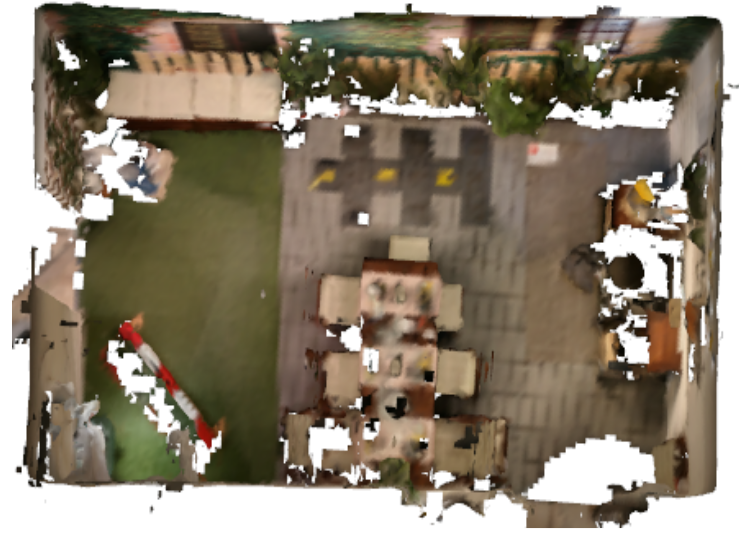}
  \vspace{-15pt}
  \caption{Monolithic no map.}
\end{subfigure}%
\begin{subfigure}{.25\textwidth}
  \centering
  \includegraphics[width=\linewidth]{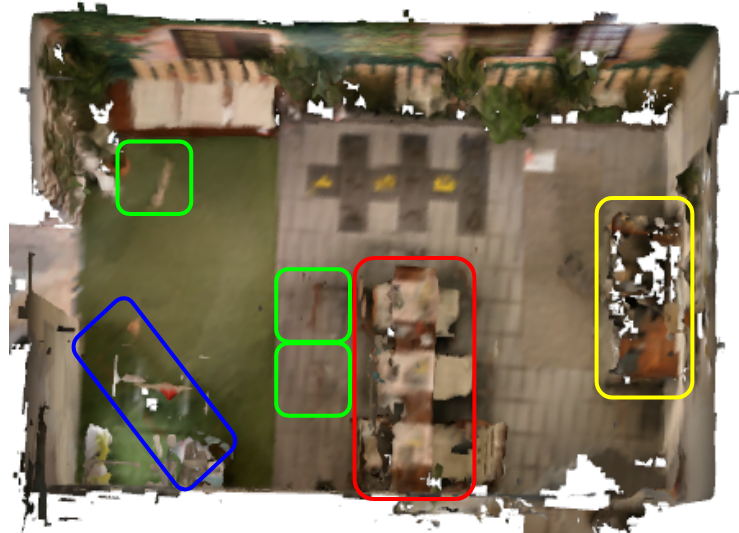}
  \vspace{-15pt}
  \caption{Monolithic with map.}
\end{subfigure}%
\begin{subfigure}{.25\textwidth}
  \centering
  \includegraphics[width=\linewidth]{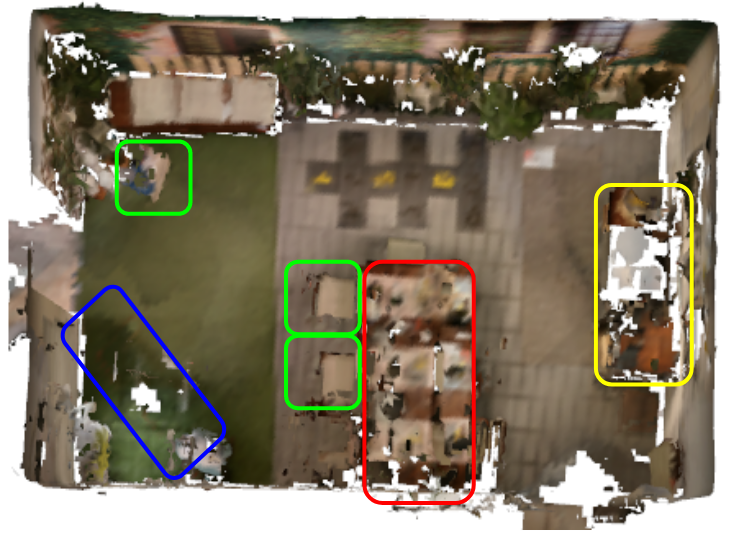}
  \vspace{-15pt}
  \caption{Ours (ground truth).}
\end{subfigure}%
\begin{subfigure}{.25\textwidth}
  \centering
  \includegraphics[width=\linewidth]{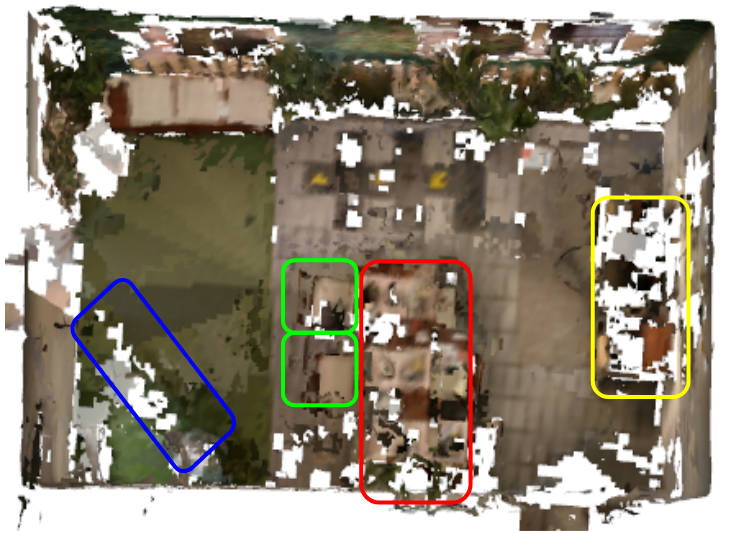}
  \vspace{-15pt}
  \caption{Ours (detectron).}
\end{subfigure}%
\vspace{-3pt}
\caption{Reconstructed meshes after run 2 in scan 466. For (c) and (d), \emph{persistent} submaps are drawn solid and \emph{unknown} submaps are shaded. Notably, (b) fails to fully reconstruct the moved table (red), which is only about half the size as in (a). 
New objects (green), such as the chairs near the table, are not captured in (b) whereas they are preserved in (c) and (d). Multiple observations at different times are merged into a blob (yellow) in (b), where our method preserves individual objects. The thin pole on (blue) is not captured by any method. The mesh in (d) appears more noisy due to the noisy Detectron2 detections. }
\label{fig:rio_qualitative}
\vspace{-10pt}
\end{figure*}

\begin{figure}[t]
    \centering
    \includegraphics[width=1.0\columnwidth]{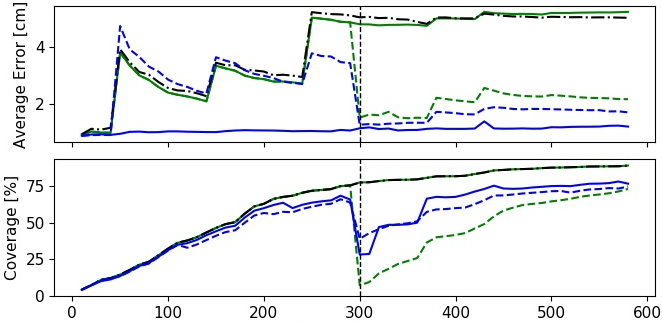}
    \includegraphics[width=1.0\columnwidth]{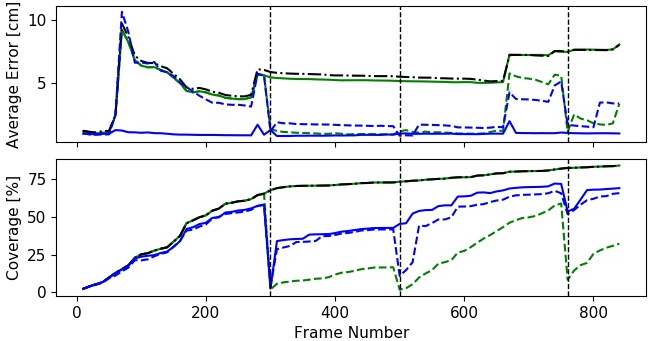}
    \includegraphics[width=1.0\columnwidth]{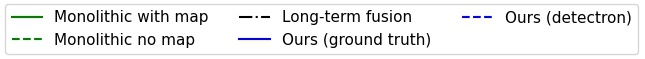}
    \vspace{-15pt}
    \caption{Adapted average error and coverage over all runs, separated by dashed lines, for scan 466 (top) and 27 (bottom).  Like in simulation, our method quickly covers the scene while keeping the error low.
    }
    \label{fig:rio_full}
        \vspace{-20pt}
\end{figure}

Experiments on the RIO dataset \cite{Wald2019RIO} verify our method on real world data. We randomly chose reference scans 466 and 27 for evaluation, where different parts of indoor scenes are reconstructed over 2 and 4 runs, respectively. To account for the increased sensor noise and the reduced Detectron2 detections, we set 
$\xi_{sdf}=2\nu$ and 
$\tau_{new}=1$. All other settings remain unchanged. Since no complete ground truth is available as in simulation, we evaluate two different approximations.


We approximate a ground truth point cloud by combining all optimized meshes of the runs. To compensate for the temporal changes, we compute the coverage as the number of ground truth points that are observed in the map, and the reconstruction accuracy as the distance from each iso-surface point to the closest ground truth point.
Fig.~\ref{fig:rio_full} shows the findings, which for both scenes are very similar. 
During the first run all errors are comparable. For our method, the reconstruction quality can be increased by eliminating high error observations that disagree with other measurements. This is particularly pronounced when using ground truth segmentation
, further highlighting the potential of leveraging semantic information for scene reconstruction. Like the simulated results, the temporal evolution after the first session demonstrates the capability of our method to represent the scene with similar or better quality than when starting from scratch, while extrapolating to a significantly larger coverage based on previous observations. The monolithic methods using the previous map show high errors and coverage. However, the coverage overestimates the true value since all points from all times are included in the ground truth. The temporal consistency of our method is further illustrated in qualitative comparisons in Fig.~\ref{fig:rio_qualitative}.

\subsection{Computational Performance}
\label{sec:res_computation}

Tab.~\ref{tab:timing} shows the mean and standard deviation of execution times per operation. Data is obtained in the second run of Sec.~\ref{sec:res_longterm}, with a sensor resolution of $640\times480$. Computation was performed on a laptop grade Intel Core i7-8550U CPU @1.80GHz.
We do not account for the panoptic segmentation, which runs at 66ms per frame according to \cite{wu2019detectron2}, although on a NVIDIA V100 GPU.

Even though our implementation is not thoroughly optimized, we achieve frame rates around 5 to 6 Hz, making our system amenable for real time operation on compute constrained mobile robots. The frame rates using real segmentation are slightly higher, since typically fewer objects are detected.
Since RIO uses a sensor resolution of $224\times172$ and fewer objects, our method speeds up significantly, highlighting the flexibility of our approach to adapt to various settings.

\begin{table}[t]
    \centering
    \caption{Computation times per operation in ms.}
      \begin{adjustbox}{max width=\columnwidth}
    \begin{tabular}{crcccc}
       Setting & Resolution & Tracking & Integration & Management & FPS$^*$ \\
    \midrule
        \multirow{2}{*}{\makecell{Flat,\\ground truth}}
        & 2-5 cm & 70.2 $\pm$ 8.4 & 104.3 $\pm$ 14.4 & 199.1 $\pm$ 54.1 & 5.1\\
        & 4-10 cm & 63.9 $\pm$ 4.4 & 89.2 $\pm$ 7.8 & 182.1 $\pm$ 44.3& 5.8\\
    \midrule
        \multirow{2}{*}{\makecell{Flat,\\detectron}}
        & 2-5 cm & 57.8 $\pm$ 5.5 & 91.1 $\pm$ 11.1 & 192.1 $\pm$ 54.3 & 5.9 \\
        & 4-10 cm & 54.7 $\pm$ 5.1 & 80.3 $\pm$ 7.4 & 183.5 $\pm$ 49.2 & 6.5 \\
    \midrule
        \multirow{2}{*}{\makecell{RIO,\\detectron}}
        & 2-5 cm & 21.8 $\pm$ 4.6 & 21.8 $\pm$ 5.9 & 33.2 $\pm$ 23.8 & 21.3
        \\
        & 4-10 cm & 16.8 $\pm$ 3.4 & 13.3 $\pm$ 3.5 & 9.5 $\pm$ 4.5 & 32.2\\
    \midrule
        \multicolumn{6}{l}{$^*$ Final frame rate is computed performing change detection every 10 frames.} \\
    \end{tabular}
    \end{adjustbox}
    \label{tab:timing}
    \vspace{-20pt}
\end{table}


\section{Conclusions}
In this work, we proposed panoptic multi-TSDFs, a novel representation for multi-resolution volumetric mapping. By leveraging higher-level information for 3D reconstruction, our proposed system allocates high resolution only where needed. Our submap-based approach achieves semantic consistency over time, enabling high reconstruction accuracy while increasing coverage by incorporating and fusing previous data where appropriate. We showed in thorough experimental validation that our map representation can be efficiently constructed, maintained, and queried during online operation on compute constrained hardware and operates robustly on real depth data and imperfect segmentation. We make our framework and data availables as open-source.

In future work, our approach can be readily combined with methods for segmentation refinement and to also account for short-term dynamics. Recognition and re-localization of changed objects could further boost performance.




{\small
\bibliographystyle{IEEEtran}
\bibliography{IEEEfull,references}

\begin{thebibliography}{10}
\providecommand{\url}[1]{#1}
\csname url@rmstyle\endcsname
\providecommand{\newblock}{\relax}
\providecommand{\bibinfo}[2]{#2}
\providecommand\BIBentrySTDinterwordspacing{\spaceskip=0pt\relax}
\providecommand\BIBentryALTinterwordstretchfactor{4}
\providecommand\BIBentryALTinterwordspacing{\spaceskip=\fontdimen2\font plus
\BIBentryALTinterwordstretchfactor\fontdimen3\font minus
  \fontdimen4\font\relax}
\providecommand\BIBforeignlanguage[2]{{%
\expandafter\ifx\csname l@#1\endcsname\relax
\typeout{** WARNING: IEEEtran.bst: No hyphenation pattern has been}%
\typeout{** loaded for the language `#1'. Using the pattern for}%
\typeout{** the default language instead.}%
\else
\language=\csname l@#1\endcsname
\fi
#2}}

\bibitem{hornung2013octomap}
A.~Hornung, K.~M. Wurm, M.~Bennewitz, C.~Stachniss, and W.~Burgard, ``Octomap:
  An efficient probabilistic 3d mapping framework based on octrees,''
  \emph{Autonomous robots}, vol.~34, no.~3, pp. 189--206, 2013.

\bibitem{oleynikova2017voxblox}
H.~Oleynikova, Z.~Taylor, M.~Fehr, R.~Siegwart, and J.~Nieto, ``Voxblox:
  Incremental 3d euclidean signed distance fields for on-board mav planning,''
  in \emph{IEEE/RSJ Int.~Conf.~on Intelligent Robots and Systems}, 2017, pp.
  1366--1373.

\bibitem{panoptic_fusion}
G.~Narita, T.~Seno, T.~Ishikawa, and Y.~Kaji, ``{PanopticFusion}: {Online}
  {Volumetric} {Semantic} {Mapping} at the {Level} of {Stuff} and {Things},''
  \emph{IEEE/RSJ Int.~Conf.~on Intelligent Robots and Systems}, pp. 4205--4212,
  Mar. 2019.

\bibitem{semantic_fusion}
J.~McCormac, A.~Handa, A.~Davison, and S.~Leutenegger, ``Semanticfusion: Dense
  3d semantic mapping with convolutional neural networks,'' in \emph{IEEE
  Int.~Conf.~on Robotics \& Automation}, 2017, pp. 4628--4635.

\bibitem{strecke_em_fusion_2019}
M.~Strecke and J.~Stueckler, ``{EM}-{Fusion}: {Dynamic} {Object}-{Level} {SLAM}
  {With} {Probabilistic} {Data} {Association},'' in \emph{IEEE/CVF
  Int.~Conf.~on Computer Vision}, Oct. 2019, pp. 5864--5873.

\bibitem{co_fusion}
M.~Rünz and L.~Agapito, ``Co-fusion: Real-time segmentation, tracking and
  fusion of multiple objects,'' in \emph{IEEE Int.~Conf.~on Robotics \&
  Automation}, 2017, pp. 4471--4478.

\bibitem{runz_maskfusion_2018}
M.~Runz, M.~Buffier, and L.~Agapito, ``{MaskFusion}: {Real}-{Time}
  {Recognition}, {Tracking} and {Reconstruction} of {Multiple} {Moving}
  {Objects},'' in \emph{{IEEE} {International} {Symposium} on {Mixed} and
  {Augmented} {Reality}}, Oct. 2018, pp. 10--20.

\bibitem{grinvald_volumetric_2019}
M.~Grinvald, F.~Furrer, T.~Novkovic, J.~J. Chung, C.~Cadena, R.~Siegwart, and
  J.~Nieto, ``Volumetric instance-aware semantic mapping and 3d object
  discovery,'' \emph{IEEE Robotics and Automation Letters}, vol.~4, no.~3, pp.
  3037--3044, 2019.

\bibitem{rosinol_kimera_2020}
A.~Rosinol, M.~Abate, Y.~Chang, and L.~Carlone, ``Kimera: {An} {Open}-{Source}
  {Library} for {Real}-{Time} {Metric}-{Semantic} {Localization} and
  {Mapping},'' in \emph{IEEE Int.~Conf.~on Robotics \& Automation}, May 2020,
  pp. 1689--1696.

\bibitem{kirillov2019panoptic}
A.~Kirillov, K.~He, R.~Girshick, C.~Rother, and P.~Doll{\'a}r, ``Panoptic
  segmentation,'' in \emph{Proc.~of IEEE Conf.~on Computer Vision and Pattern
  Recognition}, 2019, pp. 9404--9413.

\bibitem{reijgwart2019voxgraph}
V.~Reijgwart, A.~Millane, H.~Oleynikova, R.~Siegwart, C.~Cadena, and J.~Nieto,
  ``Voxgraph: Globally consistent, volumetric mapping using signed distance
  function submaps,'' \emph{IEEE Robotics and Automation Letters}, vol.~5,
  no.~1, pp. 227--234, 2019.

\bibitem{schmid2021unified}
L.~Schmid, V.~Reijgwart, L.~Ott, J.~Nieto, R.~Siegwart, and C.~Cadena, ``A
  unified approach for autonomous volumetric exploration of large scale
  environments under severe odometry drift,'' \emph{IEEE Robotics and
  Automation Letters}, vol.~6, no.~3, pp. 4504--4511, 2021.

\bibitem{xu_mid_fusion}
B.~Xu, W.~Li, D.~Tzoumanikas, M.~Bloesch, A.~Davison, and S.~Leutenegger,
  ``Mid-fusion: Octree-based object-level multi-instance dynamic slam,'' in
  \emph{IEEE Int.~Conf.~on Robotics \& Automation}, 2019, pp. 5231--5237.

\bibitem{tateno_real-time_2015}
K.~Tateno, F.~Tombari, and N.~Navab, ``Real-time and scalable incremental
  segmentation on dense {SLAM},'' in \emph{IEEE/RSJ Int.~Conf.~on Intelligent
  Robots and Systems}, vol. 2015-Decem, Dec. 2015, pp. 4465--4472.

\bibitem{whelan2015elasticfusion}
T.~Whelan, S.~Leutenegger, R.~Salas-Moreno, B.~Glocker, and A.~Davison,
  ``Elasticfusion: Dense slam without a pose graph,'' in \emph{Proc.~of
  Robotics: Science and Systems}, 2015.

\bibitem{he2017maskrcnn}
K.~He, G.~Gkioxari, P.~Doll{\'a}r, and R.~Girshick, ``Mask r-cnn,'' in
  \emph{IEEE/CVF Int.~Conf.~on Computer Vision}, 2017, pp. 2961--2969.

\bibitem{salas2013slampp}
R.~F. Salas-Moreno, R.~A. Newcombe, H.~Strasdat, P.~H. Kelly, and A.~J.
  Davison, ``Slam++: Simultaneous localisation and mapping at the level of
  objects,'' in \emph{Proc.~of IEEE Conf.~on Computer Vision and Pattern
  Recognition}, 2013, pp. 1352--1359.

\bibitem{mccormac_fusion_2018}
J.~Mccormac, R.~Clark, M.~Bloesch, A.~Davison, and S.~Leutenegger, ``Fusion++:
  {Volumetric} {Object}-{Level} {SLAM},'' in \emph{Int.~Conf.~on 3D Vision},
  Sept. 2018, pp. 32--41.

\bibitem{vespa_supereight}
E.~Vespa, N.~Nikolov, M.~Grimm, L.~Nardi, P.~H.~J. Kelly, and S.~Leutenegger,
  ``Efficient octree-based volumetric slam supporting signed-distance and
  occupancy mapping,'' \emph{IEEE Robotics and Automation Letters}, vol.~3,
  no.~2, pp. 1144--1151, 2018.

\bibitem{long_rigidfusion_2021}
R.~Long, C.~Rauch, T.~Zhang, V.~Ivan, and S.~Vijayakumar, ``{RigidFusion}:
  {Robot} {Localisation} and {Mapping} in {Environments} with {Large} {Dynamic}
  {Rigid} {Objects},'' \emph{IEEE Robotics and Automation Letters}, vol.~6,
  no.~2, pp. 3703--3710, Oct. 2021.

\bibitem{finman_2013_lifelong}
R.~Finman, T.~Whelan, M.~Kaess, and J.~J. Leonard, ``Toward lifelong object
  segmentation from change detection in dense rgb-d maps,'' in \emph{European
  Conf.~on Mobile Robots}, 2013, pp. 178--185.

\bibitem{whelan2012kintinuous}
T.~Whelan, M.~Kaess, M.~Fallon, H.~Johannsson, J.~Leonard, and J.~McDonald,
  ``Kintinuous: Spatially extended kinectfusion,'' 2012.

\bibitem{tsdf_change}
M.~Fehr, F.~Furrer, I.~Dryanovski, J.~Sturm, I.~Gilitschenski, R.~Siegwart, and
  C.~Cadena, ``Tsdf-based change detection for consistent long-term dense
  reconstruction and dynamic object discovery,'' in \emph{IEEE Int.~Conf.~on
  Robotics \& Automation}, 2017, pp. 5237--5244.

\bibitem{langer2020robust}
E.~Langer, T.~Patten, and M.~Vincze, ``Robust and efficient object change
  detection by combining global semantic information and local geometric
  verification,'' in \emph{IEEE/RSJ Int.~Conf.~on Intelligent Robots and
  Systems}, 2020, pp. 8453--8460.

\bibitem{frequency_map}
T.~Krajník, J.~Pulido~Fentanes, M.~Hanheide, and T.~Duckett, ``Persistent
  localization and life-long mapping in changing environments using the
  frequency map enhancement,'' in \emph{IEEE/RSJ Int.~Conf.~on Intelligent
  Robots and Systems}, 2016, pp. 4558--4563.

\bibitem{longterm_pointclouds}
M.~T. Lázaro, R.~Capobianco, and G.~Grisetti, ``Efficient long-term mapping in
  dynamic environments,'' in \emph{IEEE/RSJ Int.~Conf.~on Intelligent Robots
  and Systems}, 2018, pp. 153--160.

\bibitem{tang2019topological}
L.~Tang, Y.~Wang, X.~Ding, H.~Yin, R.~Xiong, and S.~Huang, ``Topological
  local-metric framework for mobile robots navigation: a long term
  perspective,'' \emph{Autonomous Robots}, vol.~43, no.~1, pp. 197--211, 2019.

\bibitem{macenski2020spatio}
S.~Macenski, D.~Tsai, and M.~Feinberg, ``Spatio-temporal voxel layer: A view on
  robot perception for the dynamic world,'' \emph{International Journal of
  Advanced Robotic Systems}, vol.~17, no.~2, p. 1729881420910530, 2020.

\bibitem{mason2012object}
J.~Mason and B.~Marthi, ``An object-based semantic world model for long-term
  change detection and semantic querying,'' in \emph{IEEE/RSJ Int.~Conf.~on
  Intelligent Robots and Systems}, 2012, pp. 3851--3858.

\bibitem{izadi2011kinectfusion}
S.~Izadi, D.~Kim, O.~Hilliges, D.~Molyneaux, R.~Newcombe, P.~Kohli, J.~Shotton,
  S.~Hodges, D.~Freeman, A.~Davison, \emph{et~al.}, ``Kinectfusion: real-time
  3d reconstruction and interaction using a moving depth camera,'' in
  \emph{Proc.~of the ACM symposium on User Interface Software and Technology},
  2011, pp. 559--568.

\bibitem{bloesch2017iterated}
M.~Bloesch, M.~Burri, S.~Omari, M.~Hutter, and R.~Siegwart, ``Iterated extended
  kalman filter based visual-inertial odometry using direct photometric
  feedback,'' \emph{Int.~Journal of Robotics Research}, vol.~36, no.~10, pp.
  1053--1072, 2017.

\bibitem{zuo2019lic}
X.~Zuo, P.~Geneva, W.~Lee, Y.~Liu, and G.~Huang, ``Lic-fusion:
  Lidar-inertial-camera odometry,'' in \emph{IEEE/RSJ Int.~Conf.~on Intelligent
  Robots and Systems}, 2019, pp. 5848--5854.

\bibitem{wu2019detectron2}
Y.~Wu, A.~Kirillov, F.~Massa, W.-Y. Lo, and R.~Girshick, ``Detectron2,''
  \url{https://github.com/facebookresearch/detectron2}, 2019.

\bibitem{lorensen1987marching}
W.~E. Lorensen and H.~E. Cline, ``Marching cubes: A high resolution 3d surface
  construction algorithm,'' \emph{ACM siggraph computer graphics}, vol.~21,
  no.~4, pp. 163--169, 1987.

\bibitem{Wald2019RIO}
J.~Wald, A.~Avetisyan, N.~Navab, F.~Tombari, and M.~Niessner, ``Rio: 3d object
  instance re-localization in changing indoor environments,'' in \emph{IEEE/CVF
  Int.~Conf.~on Computer Vision}, 2019.

\bibitem{vespa2019adaptive}
E.~Vespa, N.~Funk, P.~H. Kelly, and S.~Leutenegger, ``Adaptive-resolution
  octree-based volumetric slam,'' in \emph{Int.~Conf.~on 3D Vision}, 2019, pp.
  654--662.

\end{thebibliography}
}


\end{document}